\documentclass{article}

\usepackage[final]{neurips_2019}

\usepackage[utf8]{inputenc}
\usepackage[T1]{fontenc}
\usepackage{hyperref}
\usepackage{url}
\usepackage{booktabs}
\usepackage{amsfonts}
\usepackage{nicefrac}
\usepackage{microtype}
\usepackage{graphicx}
\usepackage{xcolor}
\usepackage{lipsum}
\usepackage{longtable}
\usepackage{amsmath} 
\usepackage{threeparttable}

\title{
  DeviceBERT: Applied Transfer Learning With Targeted Annotations and Vocabulary Enrichment to Identify Medical Device and Component Terminology in FDA Recall Summaries \\
  \vspace{1em}
}

\author{
  Miriam Farrington \\
  Department of Computer Science \\
  Stanford University \\
  \texttt{mfarring@stanford.edu} \\
}

\begin{document}

\maketitle

\begin{abstract}
\textit{FDA Medical Device recalls are critical and time-sensitive events, requiring
swift identification of impacted devices to inform the public of a recall event and
ensure patient safety. The OpenFDA device recall dataset contains valuable
information about ongoing device recall actions, but manually extracting relevant
device information from the recall action summaries is a time-consuming task.
Named Entity Recognition (NER) is a task in Natural Language Processing
(NLP) that involves identifying and categorizing named entities in unstructured text.
Existing NER models, including domain-specific models like BioBERT, struggle
to correctly identify medical device trade names, part numbers and component terms within these
summaries. To address this, we propose DeviceBERT, a medical device annotation, pre-processing and enrichment pipeline, which builds on BioBERT to identify and label medical device terminology in the device recall summaries with improved accuracy. Furthermore, we demonstrate that our approach can be applied effectively for performing entity recognition tasks where training data is limited or sparse. Source Code: \cite{project_github}}
\end{abstract}

\section{Introduction}

Large pre-trained language models have led to significant advancements in the field of biomedical text mining and entity recognition tasks. Furthermore, fine-tuned models like BioBERT have made significant strides in adapting large language models to the biomedical domain based on the original BERT architecture (\textit{\cite {devlin2019bert}}), which is trained on biomedical domain corpora from PubMed and PMC. It has been shown to achieve a significant advantage (0.62\% F1 score improvement) over BERT when applied to generalized biomedical named entity recognition tasks (\textit{\cite{Lee_2019}}). In the traditional named entity recognition, given an input sentence of \(i\) tokens \( x = \{ x_1, x_2, \ldots, x_i \} \), the NER model intends to assign each token \( x_i \) to its corresponding label \( y_i \).

\begin{figure}[!htbp]
  \centering
  \includegraphics[width=1.0\textwidth]{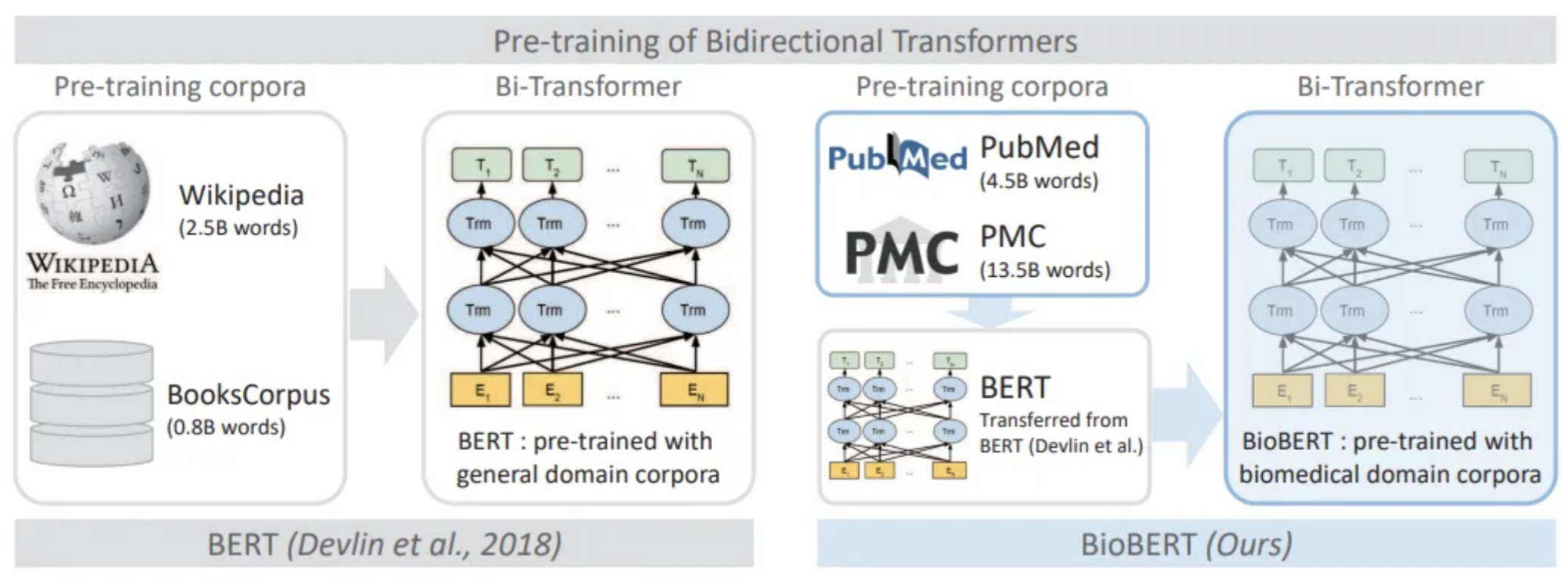}
  \caption{BioBERT Architecture \textit{\cite{Lee_2019}}}
  \label{fig:biobert-arch}
\end{figure}

Nevertheless, the performance of pre-trained models like BioBERT on domain-specific entity recognition tasks remains sub-optimal. Prior work has revealed inherent challenges when attempting to perform NER tasks in the biomedical domain, such as limited availability of training data, ambiguity in medical entity terminology, and heavy reliance on sub-domain specific acronyms in training corpora (\textit{\cite{naseem2020bioalbert}}). These shortcomings are especially apparent in the medical domain where increasingly specific terminology exists outside of the pre-trained model vocabulary.(\textit{\cite{garneau2019predicting}}). 

Prior attempts to solve for the lack of domain-specific training data have utilized pre-trained models to generate dictionaries of medical terms utilize pseudo-labeling to identify terms in unlabeled datasets; however these approaches can produce unreliable training datasets, due to the complexity of medical domain terms leading to poor generalization and improperly curated dictionary entries(\textit{\cite{wen2021medical}} Other approaches have attempted to learn out of vocabulary words through few-shot representations, with some success (\textit{\cite {hu2019fewshot}}). 

Our approach addresses these challenges through applying targeted vocabulary enrichment to BioBERT combined with a domain-specific pre-training methodology to resolve language ambiguities during tokenization and improve the quality of the available training data via consistently applied annotation rules. We call our model DeviceBERT, as it captures this methodology to demonstrate added capability to adapt to domain-specific features identifying medical device trade names, part numbers, and component terms. By tailoring BioBERT to this specific task, we aim to create a model that outperforms its general-purpose counterpart in accurately identifying medical device terminology, which can be used to better inform downstream tasks in medical device recall analysis. Additionally, we aim to show that our methodology can be applied in a generalized manner to address common shortcomings in this space and achieve satisfactory results on other domain-specific NLP tasks.

Our model outperforms BioBERT by ~13.72\% in identifying medical device, trade names, part numbers and component terms in the recall summaries. This is because our multi-phase pipeline approach optimizes BioBERT to learn domain-specific features, enhancing its ability to recognize medical device terminology. The improved performance of our model has the potential to support timely and informed decision-making in medical device recall analysis, ultimately contributing to enhanced patient safety.

\section{Related Work}
Transfer learning is a widely adopted technique in machine learning that utilizes pre-trained model weights from models trained on large-scale datasets to fine-tune the model on smaller, downstream tasks (\textit{\cite{9134370}}). This approach takes advantage of the knowledge captured by the pre-trained model on the larger dataset and adapts it to the specific requirements of the target task, thereby reducing the need for extensive retraining and improving performance.

Knowledge distillation is a related technique that involves training a smaller student model to approximate the output of a pre-trained teacher model (\textit{\cite{hinton2015distilling}}; \textit{\cite{tian2022contrastive}}, \textit{\cite{beyer2022knowledge}}). The teacher model, typically a larger and more complex model, serves as a reference for the student model, providing guidance on how to map inputs to outputs. Through this process, the student model learns to mimic the behavior of the teacher model, capturing its knowledge and expertise. The key advantage of knowledge distillation lies in its ability to retain the performance of the teacher model while significantly reducing computational requirements, making it an attractive approach for deploying models in resource-constrained environments.

\section{Approach}
Using prior work in this domain as a starting reference, we employed a multi-step approach to data retrieval, pre-processing, annotation and vocabulary enrichment in order to address some of the common shortcomings which we anticipated would present in a biomedical device entity recognition task, namely the lack of quality annotated data and anticipated pitfalls like over-fitting and poor model generalization on a limited training set. To achieve this objective, we created a pipeline to extract, curate, pre-process, tokenize a medical device dataset, then subsequently apply enrichment, regularization to create a DeviceBERT model which can be applied on downstream inferencing tasks.

\begin{figure}[!htbp]
  \centering
  \includegraphics[width=1.0\textwidth]{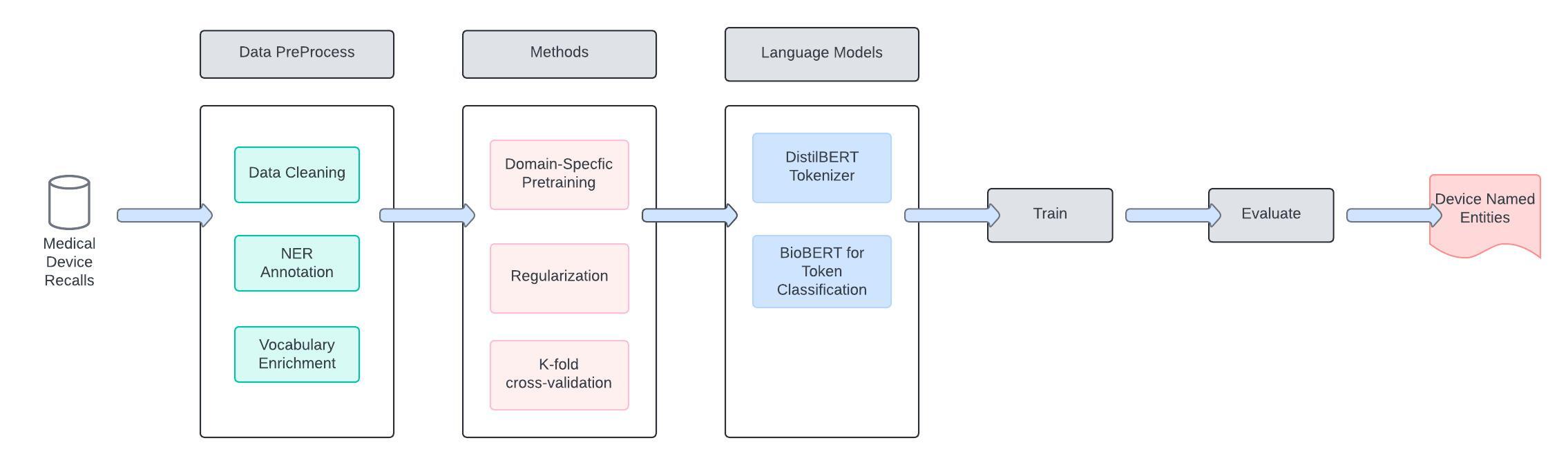}
  \caption{DeviceBERT Process Overview}
  \label{fig:devicebert-arch}
\end{figure}

First, we extracted and annotated a Medical Device NER dataset defining and applying a consistent  annotation methodology (discussed in later sections) to enable the base model to learn contextually correct, domain-specific terminology, relationships, and patterns in a recall action summary.  Pre-trained BERT tokenizers also provide sub-word tokenization, allowing for the representation of outside vocabulary words to be represented as a sequence of sub-words (\textit{\cite{wu2020sequence}}, \textit{\cite{hu2023survey}}). This is particularly useful for NER tasks, as it enables the model to capture the meaning and context of words, even when they are unseen or out-of-vocabulary (\textit{\cite{Li_2022}}). However, we are careful to observe that the BERT tokenization is not done semantically, which can lead to semantic information loss when dealing with out of vocabulary (OOV) words in domain specific downstream tasks (\textit{\cite{Nayak2020DomainAC}}). To address this problem, we enrich the vocabulary of the Distilbert tokenizer to correctly recognize and tokenize medical device words, thereby better preserving their semantic meaning. 

 To achieve this, we extract desired device vocabulary terms from the Device Recalls and Device Classification datasets, then performed a number of cleaning, normalization, de-duplication, and transformation steps before tokenizing the dataset into vocabulary words and introducing them to the DistilBERT tokenizer vocabulary. Using this approach, we identified a set of 172,800 medical device vocabulary tokens. We used this new vocabulary to increase the tokenizer vocabulary from 28,00 to 191,000 words. Following vocabulary enrichment with 100\% of new vocabulary words, the BERT tokenizer displayed significantly better recognition of medical device terms, and substantially reduced sub-word tokenization of the device recall inputs. 

\begin{table}[!htbp]
\label{tab:tokenizer_comparison}
\centering
\begin{tabular}{|p{0.45\linewidth}|p{0.45\linewidth}|}
\hline
\textbf{Before Vocabulary Enrichment} & \textbf{After Vocabulary Enrichment} \\
\hline
[CLS], An, Advisory, Letter, was, sent, to, the, customers, via, certified, mail, ., To, \#\#shi, \#\#ba, issued, a, Field, Mo, \#\#di, \#\#fication, In, \#\#struction, (, FM, \#\#I, X, \#\#RA, \#\#29, -, 90, \#\#8, \#\#28, ), to, correct, that, software, bug, and, bring, the, D, \#\#FP, -, 800, \#\#0, \#\#D, into, compliance, ., The, FM, \#\#I, is, provided, to, the, customers, at, no, charge, ., [SEP] & [CLS], An, Advisory, Letter, was, sent, to, the, customers, via, certified, mail, ., Toshiba, issued, a, Field, Modification, Instruction, (, FMI, XR, A2, 9, -, 90, \#\#8, \#\#28, ), to, correct, that, software, bug, and, bring, the, DF, P, -, 800, 0D, into, compliance, ., The, FMI, is, provided, to, the, customers, at, no, charge, ., [SEP] \\
\hline
\end{tabular}
\caption{Comparison of Tokenized Sentences Before and After Vocabulary Enrichment}
\end{table}

To utilize this data for NER, we pre-processed, tokenized, and annotated the data in a format which could be recognized by the BioBERT model to obtain a baseline, perform fine tuning, regularization, evaluation and inferencing on the device recalls dataset. BIO  (Beginning, Inside, Outside - also called IOB), is long-standing, standardized annotation format used in NER tasks to tag certain tokens in chunks of data \textit{\cite{ramshaw1995text}}. To perform BIO tagging on the tokenized recalls data, device name labels needed to be created to identify the medical device trade names and component terms in the dataset. 

Overfitting is a common risk when training language models on small downstream task datasets, leading to poor generalization performance. This occurs when the model's capacity exceeds the training data size, causing it to memorize the training data rather than learning generalized patterns(\textit{\cite {li2019overfitting}}). To mitigate the risk of overfitting,  and to ensure our model does not rely too heavily on the train/test split, we implement K-Fold cross validation on the training data, followed by dropout on the BERT embedding and encoding layers. Dropout simulates sparse activation from a given layer, encouraging the network to learn sparse representations.

\section{Device Annotation Methodology}
\begin{figure}[!htbp]
  \centering
  \includegraphics[width=1.0\textwidth]{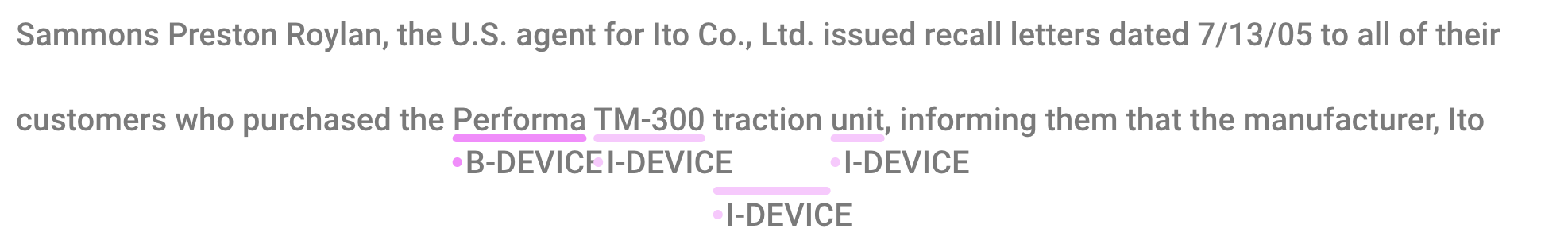}
  \caption{Partial text example of an annotated recall action with BIO tagging applied to a device trade name}
  \label{fig:annotation}
\end{figure}

We performed a number of pre-processing steps which involved extracting the recall action summary text from the larger dataset with the unique identifier (cfres\_id) and then performed random sampling across the dataset to extract a representative subset of recalls data for annotation (Figure~\ref{fig:annotation}). To perform the annotations of the recall data we utilized the doccano open source tool, which provides a web server interface and mysql database for defining and applying annotations \cite{doccano}. After processing the input schema into a format that can be understood by doccano, we defined 3 custom NER labels (B-DEVICE, I-DEVICE, O-DEVICE) to apply to the data. In order to ensure consistency in the application of NER labels, the following labeling methodology was used: 

\begin{itemize}
    \item The beginning word of a device name, component part or part number is assigned the label B-DEVICE 
    \item Subsequent words in the device name which should be given attention are assigned the label I-DEVICE
    \item Outside words which are part of the device name ('and', 'the', etc) which should be excluded from attention are assigned the label O-DEVICE 
    \item The remaining tokens are assigned the value of 'O', which indicates they are not tokens which are part of any Device terms. 
    \item Software components of a medical device are excluded from NER labeling, which includes software trade names, version numbers, and operating systems.
    \item Different Device model numbers/names are treated as separate devices. 
    \item Context of words is considered when applying NER labels; if a word has multiple meanings in different contexts, the NER label is only applied where the term is used to refer to a medical device.
    \item Special characters included in the device word or between device words were excluded from NER labels.
\end{itemize}

Using this method, we annotated 10\% of the recall actions source data after de-duplication; which produced a labeled dataset of just over 2000 unique recall actions. During annotation, we observed that the device label distribution is largely sparse, meaning the NER labels are applied infrequently and, occasionally, not at all within a given recall action.

Once annotation was completed, the recalls text is extracted from doccano and tokenized into words. The NER label tags are created and mapped to each word token at the positions designated in the original label span. A 2-directional label mapping is then created mapping each label to its numeric representation (0-3, with a special tag designated for unlabeled tokens (-100)). This ensures that only the tokens which are part of named entities contribute to the loss. This effectively masks the remaining tokens and ensures that they do not influence the gradients during back-propagation.

The training and evaluation process closely follows the HuggingFace Transformers documentation for some key implementation steps and code \cite {huggingface-token-class}. In this case, we chose to create a custom dataset based on the previously annotated recalls data and then initialize a DistilBERT tokenizer and BioBERT base model for token classification utilizing the HuggingFace transformers library, which is supported by the original BioBERT project \cite{biobert-base}. This allows for flexibility in experimentation, utilizing permutations of the original input data converted to the Huggingface Datset class. 

To perform padding of the inputs and labels, we use a DataCollatorForTokenClassification, which performs dynamic padding of the tokens as they are processed, selecting the max\_length argument.  Next, using the labeled dataset in a train-test split, we fine-tune the BioBERT model on the recalls dataset to recognize medical device terms and evaluate its performance to establish a baseline. We will experimented with additional hyper-parameter optimization methods to further improve the model performance and record the results as f1 scores. Finally,we compare the results from each experiment and report our results, along with suggestions for further optimization efforts and potential other expansions or applications of our findings to new research areas.

\section{Experiments}
To establish a baseline metric for the evaluation of DeviceBERT, we trained a base BERT model and BioBERT model on the corpus of device recall actions. Our hypothesis was that BioBERT would offer at the least a marginal advantage over BERT, due to the pre-trained medical domain knowledge incorporated into BioBERT. 

Next, we conduct several experiments to evaluate the performance of DeviceBERT: 

\begin{itemize}
    \item DeviceBERT utilizing combined regularization techniques on the DistilBERT base tokenizer (+Reg only). 
    \item DeviceBERT utilizing DistilBERT tokenizer with medical device vocabulary enrichment (100\% enrichment) 
    \item DeviceBERT utilizing DistilBERT tokenizer with medical device vocabulary enrichment (50\% enrichment) 
    \item DeviceBERT utilizing DistilBERT tokenizer with medical device vocabulary enrichment (25\% enrichment) 
    \item DeviceBERT with a cimbination of regularization techniques and medical device vocabulary enrichment (+Reg+Vocab). 
\end{itemize}

Below is a brief summary of the dataset evaluation and experiments conducted, which are reported in Table \ref{tab:ner_comparison}).

\subsection{Data}
The FDA Center for Devices and Radiological Heath (CDRH) makes available one of the most comprehensive public datasets of medical device names, terms, product numbers and recall actions \textit{\cite{openfda_recalls}}. This work relies on datasets made available through the Open FDA API. The CDRH Device Recalls Dataset contains 20,150 unique recall actions (see Figure \ref{tab:recall_action_sample}). We exclude duplicate recall description records and records which did not contain an associated recall action from the dataset. 

\begin{table}[h]
\centering
\begin{tabular}{|c|p{10cm}|}
    \hline
    \textbf{recall\_id} & \textbf{recall\_action} \\ \hline
    207218 & QuidelOrtho issued Important Product Correction Notification issued April 9, 2024. Letter states reason for recall, health risk and action to take: \\
           & Discontinue using, render unusable, and discard affected lots of VITROS Free T3 Reagent Pack (and associated calibrators). \\
           & Complete the enclosed Confirmation of Receipt form no later than April 17, 2024. Upon receipt of your completed form, QuidelOrtho will provide credit for, or replacement of, discarded product. \\
           & Save this notification with your User Documentation or post this notification by each VITROS ECi/ECiQ/3600/5600/XT 7600 System until the issue has been resolved. \\
           & Please forward this notification if the affected product was distributed outside of your facility. \\
           & If your laboratory has experienced the issue with this product and you have not already done so, please report the occurrence to your local Global Services Organization. \\
           & Resolution: QuidelOrtho's investigation is ongoing and currently working to identify root cause. \\
           & Contact Global Services Organization at 1-800-421-3311. \\ \hline
\end{tabular}
\caption{Sample Recall Action}
\label{tab:recall_action_sample}
\end{table}

To compile the tokenizer vocabulary, we extract the device\_name and product\_description fields from the Device Recalls dataset, and combine with the FDA Device Registration database. The data is cleaned to remove certain characters and tokens, de-duplicated, shuffled and tokenized into words. Tokens which consist of purely numeric values were excluded from the vocabulary. This reduces the total tokens from 4,034,870 to 172,821 tokens added to the tokenizer vocabulary, increasing the size from 28,996 to 191,049 when 100\% of the identified tokens were added to the vocabulary. 

We conducted several experiments augmenting the tokenizer vocabulary using smaller vocabulary token subset splits of  50\%, 25\%. We also conducted additional experiments using various methods of token batch splitting (not reported here) by modifying the input order of the tokens, using random shuffle, First in First Out (FIFO) and sorting by token length in descending order. However, much like prior work in this area, we found that the manner in which the tokens were added to the vocabulary did not substantially impact the results of training on the split subset, so long as the tokens themselves were sufficiently representative of the sub-domain device vocabulary. (\textit{\cite{wei2022exploring}})

\subsection{Evaluation method}
Precision, Recall and F1 score are used as standard metrics to evaluate the performance of the model on each NER task. For consistency in reporting the results of our experiments, we use the same metrics used to evaluate the original BioBERT model on a given dataset (\textit{\cite{Lee_2019}}). 

\subsection{Experimental details}
Adam Optimizer is a popular adaptive gradient descent method used in deep learning language models. However, it is known to perform poorly on generalization using smaller datasets (\textit{\cite{zou2021understanding}}). For this reason, we  use the modified AdamW optimizer when conducting training experiments, which utilizes an adaptive, decoupled weight decay, whereby the regularization term is only proportional to the weight itself.(\textit{\cite{loshchilov2019decoupled}}). The terms can be expressed as: 

$$
\begin{aligned}
m_t &= \beta_1 \cdot m_{t-1} + (1 - \beta_1) \cdot \nabla \theta_t \\
v_t &= \beta_2 \cdot v_{t-1} + (1 - \beta_2) \cdot \nabla \theta_t^2 \\
\hat{m}_t &= \frac{m_t}{1 - \beta_1^t} \\
\hat{v}_t &= \frac{v_t}{1 - \beta_2^t} \\
\theta_{t+1} &= \theta_t - \alpha \cdot \frac{\hat{m}_t}{\sqrt{\hat{v}_t} + \epsilon} - \alpha \cdot w \cdot \theta_t
\end{aligned}
$$

For our experiments, we initialized the inputs as follows: 

\[
\text{Learning Rate } (\alpha): 1 \times 10^{-5}
\]
\[
\text{Beta 1 } (\beta_1): 0.9
\]
\[
\text{Beta 2 } (\beta_2): 0.999
\]
\[
\text{Epsilon } (\varepsilon): 1 \times 10^{-8}
\]
\[
\text{Weight Decay } (w): 0.02
\]

All experiments reported here were run using 1 L4 NVIDIA GPU. The batch size and input length are set to 16 and 512, respectively.

\subsection{Results}
The results of the NER Tests conducted are reported in Table \ref{tab:ner_comparison}. First, we observe that in our baseline testing BioBERT, which was trained on the biomedical domain corpus, performs less optimally than anticipated, indicating that it was not able to generalize well to the sub-domain of medical device terminology. Meanwhile, BERT, which was trained on only the general knowledge corpus, actually slightly outperforms BioBERT in Recall and F1 score, however the margin is so narrow that it cannot be said one model substantially outperformed the other.

\begin{table}[!htbp]
\caption{Performance comparison of language models on NER dataset of devices}
\label{tab:ner_comparison}
\centering
\begin{threeparttable}
\begin{tabular}{lccc}
\toprule
\textbf{Model} & \textbf{Precision (\%)} & \textbf{Recall (\%)} & \textbf{F1 Score (\%)} \\
\midrule
BERT & 72.96 & 73.99 & 73.47 \\
BioBERT & 73.42 & 73.29 & 73.35 \\
\midrule
\multicolumn{4}{c}{\textbf{DeviceBERT}} \\
\midrule
(+Reg only) & 82.37 & 78.52 & 80.37 \\
(+Vocab 100\%) & 75.59 & 73.46 & 74.51 \\
(+Vocab 50\%) & 81.56 & 80.11 & 80.83 \\
(+Vocab 25\%) & 80.14 & 77.87 & 78.91 \\
(+Reg+Vocab) & 85.14 & 82.07 & 83.56 \\
\bottomrule
\end{tabular}
\begin{minipage}{\linewidth}
\small
\begin{flushleft}
\textbf{Note:} Best scores are reported from all epochs. Precision, recall, and F1 scores for BERT and BioBERT base cased models trained on the sub-domain device recalls text are provided as a baseline metric. DeviceBERT scores are displayed as a computed average of cross-validation scores.
\end{flushleft}
\end{minipage}
\end{threeparttable}
\end{table}

Results for DeviceBERT across all experiments show that the combination of pre-training, coupled with vocabulary enrichment achieves the highest overall F1 score. One interesting observation is the relative high performance of DeviceBERT utilizing only regularization techniques, which we feel warrants further investigation, as the use of regularization techniques alone appear to correlate with a significant impact on overall model accuracy across all 3 metrics.

\section{Analysis}
As shown in the vocabulary enrichment results, the DeviceBERT model performs most effectively on a vocab token split of 50\% which provides a balanced tradeoff between sufficient vocabulary to tokenize the dataset effectively, without succumbing to the phenomenon of catastrophic forgetting, which can occur when the model is introduced to a large set of of new data (\textit{\cite{cossu2022forgetting}}). This split was chosen and combined with regularization techniques to evaluate the combined score (Reg+Vocab), where DeviceBERT performed best. 

\begin{figure}[!htbp]
  \centering
  \includegraphics[width=0.7\textwidth]{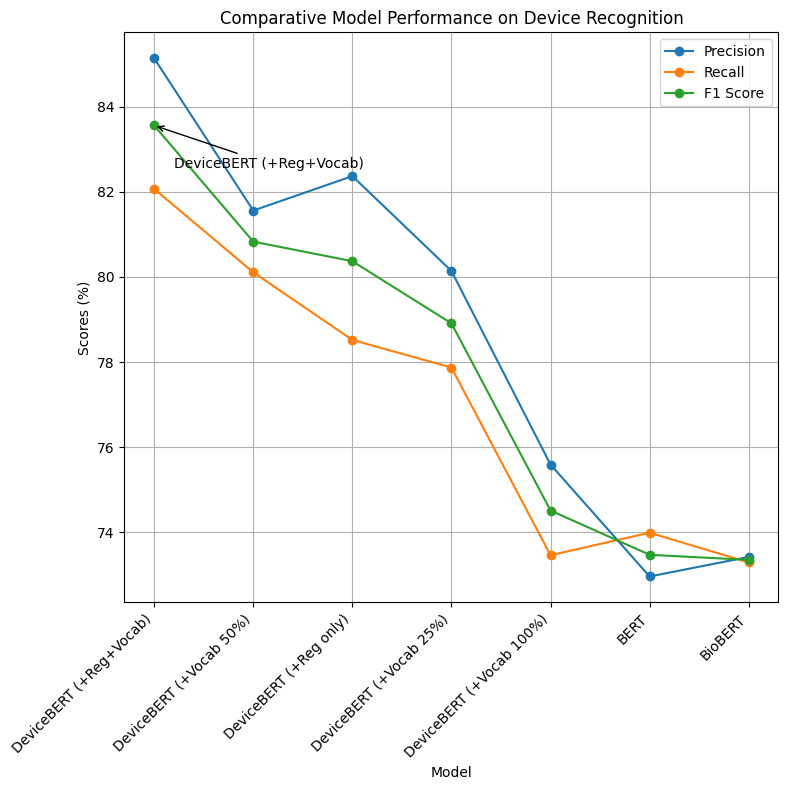}
  \caption{Comparative score of all models on device entity recognition task.}
  \label{fig:model-plot-results}
\end{figure}

The comparative analysis shown in Figure~\ref{fig:model-plot-results} points to a significant improvement in F1 score bewteen our baseline BERT models and DeviceBERT. Using the combined approach outlined, we were able to achieve an approximately ~13.72\% improvement in overall F1 score.

Our analysis finds that DeviceBERT performs best when utilizing a combination of pre-training steps and vocabulary enrichment coupled with standard regularization techniques to help mitgate overfitting.

\section{Conclusion}
DeviceBERT is proposed to maximize the use of an elaborately pre-trained domain-specific model (BioBERT) for performing certain biomedical sub-domain tasks, by enriching a pre-trained tokenizer's vocabulary,  thereby improve semantic retention during tokenization. DeviceBERT adds a new domain-specific medical device vocabulary, while using an ensemble of annotation, cross-validation and regularization techniques to avoid some of the pitfalls leading to overfitting when working with limited training data. 

The DeviceBERT approach significant improves the efficiency of modifying a base pre-training model like BERT or BioBERT for a new
sub-domain. With DeviceBERT we can leverage the power of transfer learning using pre-trained language models for these new sub-domains without being hindered by limited training data. The approach could be particularly attractive to ad-hoc and other special-purpose domains. DeviceBERT also paves the way for potential new applications of the model to downstream Named Entity Linking tasks. 

\section{Future work}

While the results reported are promising, more work can be done to improve the performance of DeviceBERT on device recognition. We recommend a more robust training dataset using the existing annotation methodology  as a potential avenue for future work in addition to curating device data from alternative sources that could bring greater contextual and semantic understanding. Finally, as regularization showed promising results, we hypothesize that the experimentation using additional regularization techniques could further boost overall performance on the task.

Additionally, the techniques and methodology proposed for the implementation of DeviceBERT lays the groundwork for future research in the application of Device Entity Recognition to Device Named Entity Linking or NEL. NEL is a growing field in Artificial Intelligence research which involves a threefold process of entity recognition, disambiguation, and linking (\textit{\cite {tedeschi-etal-2021-named-entity}}) Generic device terms ("kit","unit", "graph"), etc can have multiple meanings. By applying the NER methodology described here to the task of NEL, future work could attempt to tie the named entities from the recall actions to their corresponding unique identifier, in order to disambiguate the more generic terms and provide greater context and deeper natural language understanding of the semantic meaning of medical device terms.

\section{Ethics Statement}
A critical consideration when working with models trained on biomedical datasets is to always ensure that the training data which utilizes patient information is ethically sourced and respects patient privacy. The experiments conducted for this task utilized publicly available recalls datasets that do not contain Personally Identifiable Information (PII); however that information could be easily introduced in a downstream training or fine-tuning task. Before including any PII data in training, steps should be taken to obtain patient informed consent where applicable and then ensure the data is properly anonymized and masked to prevent identification. 

\bibliographystyle{plainnat}
\bibliography{references}

\appendix

\section{Appendix}

\begin{figure}[!htbp]
  \centering
  \includegraphics[width=1.0\textwidth]{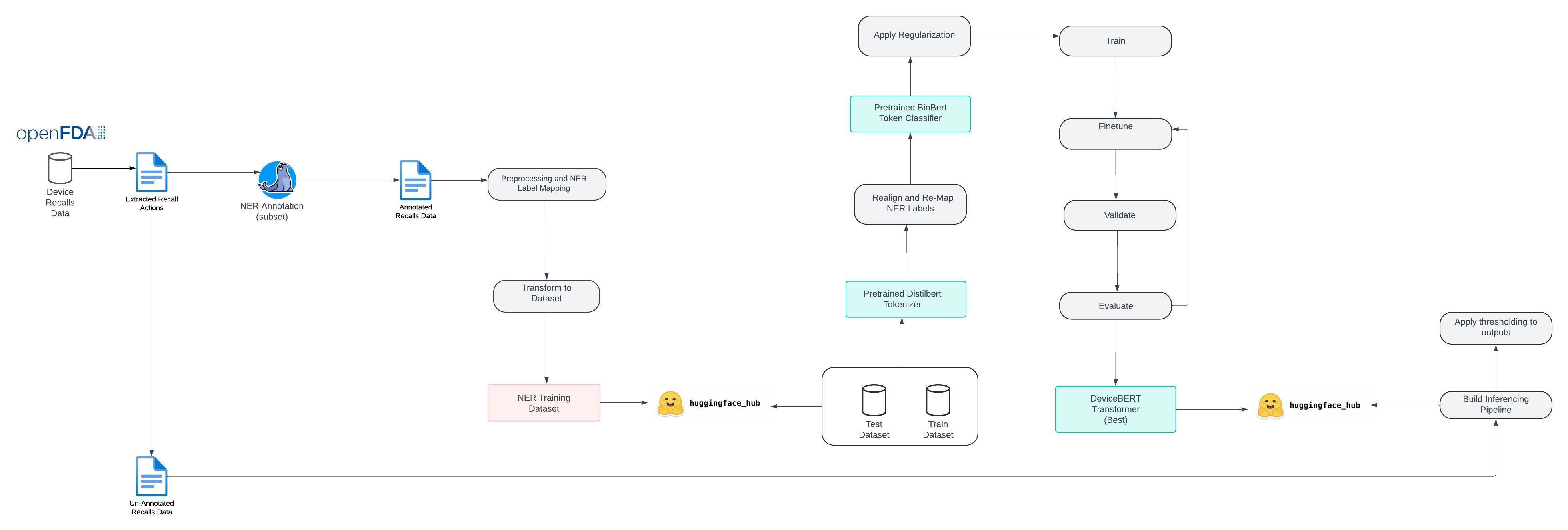}
  \caption{DeviceBERT Process Flow Diagram}
  \label{fig:biobert-ref-arch}
\end{figure}

We include additional diagrams that may be useful to understanding the full data pre-training and process flow in training and evaluating DeviceBERT (See Figure~\ref{fig:biobert-ref-arch}). This diagram breaks down the steps involved in the complete pipeline from data extraction and curation, to pre-processing, enrichment, validatation, regularization, training, evaluation and finally inferencing. 

Additionally, for reference, we include an example output from the various inferencing experiments that were performed utilizing DeviceBERT. The table lists the original device recall text in full followed by the predicted NER Device labels for tokens. To make the task more easy to analyze, we set a threshold of 0.99 for the predicted label, and eliminated token predictions which fell below that threshold. The result is shown in the table, where all the listed tokens represent words with a label prediction of >= 0.99. (See Table \ref{tab:recall}).

\vspace{12cm}

\title{NER Labels for Recall Text}
\author{}
\date{}
\maketitle
\begin{longtable}{|p{7cm}|p{7cm}|}
\hline
\textbf{Recall Text} & \textbf{Label} \\
\hline
\begin{minipage}[h]{\linewidth}
\small
\textit{Philips Healthcare sent an URGENT-MEDICAL DEVICE RECALL letter dated October 15, 2012 to all affected customers. The letter identified the product, problem, and actions to be taken by the customers. Customers were instructed to inspect all casters of the unit to ensure that they are all secured. If a caster is loose, customers were told to lock the caster in place, limit movement of the cart and contact their local Phillips Invivo Representative. Nevertheless, a Philips Invivo representative will contact the customer regarding their affected device. All affected devices will have new casters installed in order to correct the problem. Contact your local Philips Invivo Representative at 1-800-722-9377 for further information and support.}
\end{minipage}
&
\begin{minipage}[h]{\linewidth}
\small
Token: DEVICE - Label: I-DEVICE \\
Token: RECALL - Label: I-DEVICE \\
Token: product - Label: B-DEVICE \\
Token: c - Label: B-DEVICE \\
Token: casters - Label: B-DEVICE \\
Token: caster - Label: B-DEVICE \\
Token: caster - Label: B-DEVICE \\
Token: Invivo - Label: I-DEVICE \\
Token: Representative - Label: I-DEVICE \\
Token: . - Label: I-DEVICE \\
Token: representative - Label: I-DEVICE \\
Token: customer - Label: B-DEVICE \\
Token: device - Label: B-DEVICE \\
Token: RECALL - Label: I-DEVICE \\
Token: product - Label: B-DEVICE \\
Token: customers - Label: B-DEVICE \\
Token: casters - Label: B-DEVICE \\
Token: caster - Label: B-DEVICE \\
Token: caster - Label: B-DEVICE \\
Token: cart - Label: B-DEVICE \\
Token: Invivo - Label: I-DEVICE \\
Token: Representative - Label: I-DEVICE \\
Token: . - Label: I-DEVICE \\
Token: Philips - Label: B-DEVICE \\
Token: representative - Label: I-DEVICE \\
Token: customer - Label: B-DEVICE \\
Token: device - Label: B-DEVICE \\
Token: devices - Label: B-DEVICE \\
Token: casters - Label: B-DEVICE \\
Token: installed - Label: B-DEVICE \\
Token: problem - Label: B-DEVICE \\
Token: Representative - Label: I-DEVICE \\
\end{minipage}
\\
\hline
\begin{minipage}[t]{\linewidth}
\small
\textit{The firm disseminated a Medical Device Safety Alert by letter beginning on 04/03/2020. The notice explained the potential for damage that can occur during the procedure. Specifically, the Outflow Graft may be subject to tears and the Strain Relief screw may break during the pre-implant pump assembly and attachment to the HVAD Pump. The letter also provided additional steps for assembly and attachment to reduce the risk of damage and tearing during the assembly procedure which was provided in Appendix A.}
\end{minipage}
&
\begin{minipage}[t]{\linewidth}
\small
Token: Medical - Label: B-DEVICE \\
Token: Device - Label: I-DEVICE \\
Token: Safety - Label: I-DEVICE \\
Token: Alert - Label: I-DEVICE \\
Token: by - Label: I-DEVICE \\
Token: letter - Label: I-DEVICE \\
Token: notice - Label: B-DEVICE \\
Token: procedure - Label: B-DEVICE \\
Token: Outflow - Label: B-DEVICE \\
Token: Graft - Label: I-DEVICE \\
Token: may - Label: I-DEVICE \\
Token: be - Label: I-DEVICE \\
Token: subject - Label: I-DEVICE \\
Token: to - Label: I-DEVICE \\
Token: Strain - Label: B-DEVICE \\
Token: Relief - Label: I-DEVICE \\
Token: screw - Label: I-DEVICE \\
Token: may - Label: I-DEVICE \\
Token: break - Label: I-DEVICE \\
Token: pump - Label: I-DEVICE \\
Token: assembly - Label: I-DEVICE \\
Token: HVAD - Label: B-DEVICE \\
Token: Pump - Label: I-DEVICE \\
Token: . - Label: I-DEVICE \\
Token: assembly - Label: B-DEVICE \\
Token: and - Label: B-DEVICE \\
Token: damage - Label: B-DEVICE \\
Token: assembly - Label: B-DEVICE \\
\end{minipage}
\\
\hline
\caption{Inferencing Result with Thresholding on Device Recall Actions} \\
\label{tab:recall}
\end{longtable}
\end{document}